\documentclass{article}

\usepackage{arxiv}

\usepackage[utf8]{inputenc} 
\usepackage[T1]{fontenc}    
\usepackage{hyperref}       
\usepackage{url}            
\usepackage{booktabs}       
\usepackage{amsfonts}       
\usepackage{nicefrac}       
\usepackage{microtype}      
\usepackage{lipsum}		
\usepackage{graphicx}
\usepackage[round]{natbib}  
\usepackage{doi}

\title{DuetRAG: Collaborative Retrieval-Augmented Generation}

\author{Dian Jiao \\
        Zhejiang University\\
	\texttt{jd\_dcd@zju.edu.cn} \\
	\And
        Li Cai \\
	Zhejiang University\\
	\texttt{22321284@zju.edu.cn} \\
    \And
        Jingsheng Huang \\
	Zhejiang University\\
	\texttt{jingsheng@zju.edu.cn} \\
    \And
        Wenqiao Zhang \\
	Zhejiang University\\
	\texttt{wenqiaozhang@zju.edu.cn} \\
     \And
        Siliang Tang \\
	Zhejiang University\\
	\texttt{siliang@zju.edu.cn} \\
    \And
        Yueting Zhuang \\
	Zhejiang University\\
	\texttt{yzhuang@zju.edu.cn} \\
}

\renewcommand{\shorttitle}{DuetRAG: Collaborative Retrieval-Augmented Generation}
\newcommand*{\method}{DuetRAG}
\hypersetup{
pdftitle={A template for the arxiv style},
pdfsubject={q-bio.NC, q-bio.QM},
pdfauthor={Dian Jiao, Li Cai},
pdfkeywords={First keyword, Second keyword, More},
}

\begin{document}
\maketitle

\section{Abstract}
Retrieval-Augmented Generation (RAG) methods augment the input of Large Language Models (LLMs) with relevant retrieved passages, reducing factual errors in knowledge-intensive tasks. However, contemporary RAG approaches suffer from irrelevant knowledge retrieval issues in complex domain questions (\emph{e.g.}, HotPot QA) due to the lack of corresponding domain knowledge, leading to low-quality generations. To address this issue, we propose a novel  Collaborative Retrieval-Augmented Generation framework, \method{}. Our
bootstrapping philosophy is to simultaneously integrate the domain fintuning and RAG models to improve the knowledge retrieval quality, thereby enhancing generation quality.
Finally, we demonstrate \method{}’s matches with expert human researchers on HotPot QA.

\section{Introduction}
Large language models (LLMs) have demonstrated significantly enhanced performance compared to preceding efforts within the realm of question answering (QA) \citep{achiam2023gpt,wei2022chain}. Nonetheless, in the domain of domain-specific knowledge QA, LLMs frequently exhibit subpar performance attributable to an insufficiency of pertinent knowledge. LLMs, when fine-tuned on domain-specific knowledge, acquire specialized domain knowledge to address queries via fine-tuning on an extensive corpus of domain-specific information. However, this methodology typically necessitates copious amounts of data and entails substantial training overhead, while concurrently exposing the model to the risk of catastrophic forgetting \citep{kirkpatrick2017overcoming}, thereby impeding its ability to achieve desired outcomes in QA across alternate domains. The paradigm of question answering grounded in retrieval-augmented generation (RAG) facilitates the integration of external knowledge into LLMs \citep{lewis2020retrieval}, thereby enabling them to extract contextually pertinent information from external repositories in instances where the model lacks internal knowledge, thus improving the performance of large language models in domain-specific knowledge question answering. However, traditional RAG suffers from the illusion problem \citep{zhang2023siren}, leading to incorrect answers, and when faced with more complex questions, limited by the performance of the retriever, RAG may encounter difficulties in answering based on poorly correlated or even erroneous external documents.

In this paper, we propose a novel Collaborative Retrieval-Augmented Generation framework, called DuetRAG, based on RAG and domain knowledge fine-tuning. This framework leverages both fine-tuned models and RAG to generate answers to questions, and a referee model is employed to determine the selection of the final answer. The aim of DuetRAG is not only to equip models with domain-specific knowledge but also to enable them to utilize external documents to obtain answers when internal knowledge is uncertain. By complementing internal and external knowledge, DuetRAG aims to enhance the robustness of the model. We provide a detailed description of the DuetRAG methodology and its performance on various datasets in Section 3. DuetRAG outperforms existing RAG and fine-tuned models on the HotPot QA dataset \citep{yang2018hotpotqa}, presenting a new pipeline for QA.

\section{Related Work}

\textbf{Retrieval-Augmented Language Models} Augmenting language models with relevant information obtained from various external knowledge bases has been shown to significantly improve the performance of various NLP tasks, including language modeling \citep{guu2020retrieval,borgeaud2022improving,shi2023replug,lin2023ra} and open domain question answering \citep{izacard2022atlas,zhang2024raft}.
RAG mainly adopts the "retrieve then read" paradigm. Specifically, the input question is first used as the query, then the retrieval module retrieves relevant documents from the external knowledge base, and finally the retrieved documents and questions are merged into a complete input to generate final output.
For example, RETRO \citep{borgeaud2022improving} modifies the autoregressive LM to focus on relevant documents through chunked cross-attention, thereby introducing new parameters to the model. REPLUG \citep{shi2023replug} assumes black-box access to LM and optimizes it by fine-tuning the retriever. RAFT \citep{zhang2024raft} proposes a fine-tuned data that additionally contains relevant documents and answers with reasoning chains to train language models for domain-specific open-book settings.\\
\textbf{Finetuning for RAG} Recently, related work has studied how to improve the overall performance by fine-tuning the LLM or retriever in the RAG framework. For example, RADIT \citep{lin2023ra} proposes a dual-instruction fine-tuning framework to fine-tune both the LLM and the retriever simultaneously. InstructRetro \citep{wang2023instructretro} pre-trains a larger autoregressive large-scale language model with retrieval function and performs instruction fine-tuning based on it. ChatQA \citep{liu2024chatqa} additionally proposes a context-enhanced instruction fine-tuning stage, specifically to enhance the model's ability to perform context awareness in conversational QA. RAFT 
\citep{zhang2024raft} proposes a kind of fine-tuning data that additionally contains related documents and answers with reasoning chains to fine-tune LLM and improve LLM's ability to understand the retrieved documents under the RAG framework.\\
\textbf{Multi-Model collaboration} Benefiting from the extensive knowledge exhibited by large language models, the use of multi-model collaboration to solve or improve tasks is gradually being studied. CAMEL \citep{li2024camel} proposes a cooperative agent framework called role-playing that enables agents to cooperate on complex tasks. \citet{du2023improving} and \citet{liang2023encouraging} explore enhancing the factuality of specific tasks, e.g., translation and arithmetic reasoning, by facilitating “debates” among multiple models. Corex \citep{sun2023corex} has explored three collaboration paradigms in reasoning – debate, code review and retrieve. In addition, model collaboration can also be used between models of different sizes. \citet{li2024guiding} generates prompts through a small model to guide the generation of LLM.



\begin{figure}[htbp]
  \centering
  \includegraphics[width=0.9\textwidth, trim=0.02in 0.02in 0.02in 0.02in, clip]{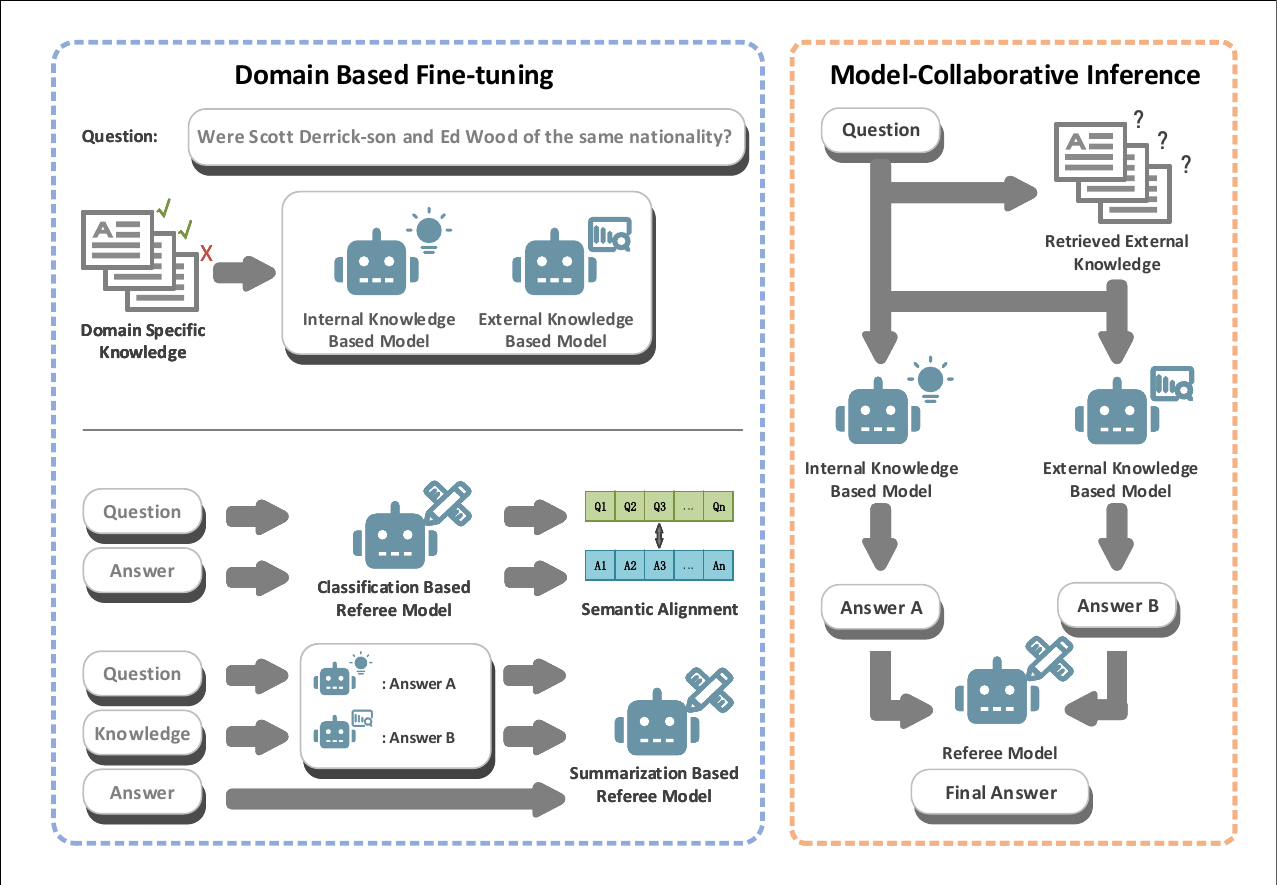}
  \caption{\textbf{Overview of our DUET method.}}
  \label{fig:main}
\end{figure}

\section{DuetRAG: Collaborative Retrieval-Augmented Generation}
In this section, we introduce DuetRAG, a novel model-collaborative approach for retrieval-based argument generation, as shown in Figure 1. The DuetRAG framework relies on the synergy of three models: the Reciter, which utilizes fine-tuning to internalize a knowledge base; the Discoverer, which employs retrieval and refinement to obtain external knowledge; and the Arbiter, which evaluates and summarizes the answers from the first two models. We will first present the Problem Formalization, followed by a detailed introduction to our proposed DuetRAG.

\subsection{Problem Formalization}
Formally, for a specific domain task T, we have access to a training set ${Train\{Q_{train}, D_{train}, A_{train}\}}$,  Here, $Q_{train}$ represents the questions in the training set, $D_{train}$ is a collection of specific documents that contain the answers to the questions in the training set, and $A_{train}$ is the correct answers to the questions $Q_{train}$. During the testing phase, we have access to a test set $Test\{Q_{test},  D\}$, where $Q_{test}$ is the set of questions from the test task, and D is the external knowledge collection. For the training set ${Train\{Q_{train}, D_{train}, A_{train}\}}$, $D_{train}$ can be further classified into documents $d^*_1, d^*_2, ..., d^*_i$ that contain direct and relevant information to the answers, and noise documents $d_1, d_2, ..., d_j$ that are irrelevant to the answers.

\subsection{Overview of DuetRAG}
The core of our work, DuetRAG, involves the collaboration of three models: the question-answering model $\mathcal{M}_i$, which utilizes internal knowledge through fine-tuning based on domain expertise to answer questions; the question-answering model $\mathcal{M}_e$, which utilizes external knowledge by fine-tuning based on domain knowledge and question-answering scenarios to answer questions; and the referee model $\mathcal{M}_j$, which evaluates the answers from the aforementioned question-answering models and selects the answer with the highest confidence.

During the Fine-tuning Stage, $\mathcal{M}_i$ learns domain-specific knowledge and recalls relevant knowledge within the model, while $\mathcal{M}_e$ learns how to retrieve and summarize answers from external documents. $\mathcal{M}_j$ is tasked with assessing the credibility of answers and generating the highest-confidence result based on existing answers.

\subsection{Fine-tuning with Domain Knowledge}
Our work, DuetRAG, employs a domain knowledge fine-tuning approach based on LoRA to optimize the Internal Knowledge Based Model $\mathcal{M}_i$ and the External Knowledge Based Model $\mathcal{M}_e$. For a specific domain T and training set $T_{train}$, we combine domain questions $Q_train$, domain knowledge $D_{train}$, and question answers $A_{train}$ into a prompt ${q, d, a}$ used for LoRA fine-tuning. The weights of the large model contain a vast amount of knowledge learned during the training phase, and through domain knowledge fine-tuning, the model's knowledge relevant to the target domain will be reactivated. At the same time, the large language model possesses reasoning and summarization capabilities. Through domain knowledge fine-tuning, the model can learn to extract and answer questions related to external knowledge.

Our work on the referee model $\mathcal{M}_j$, responsible for assessing the credibility of answers and generating the final result, involves the design and implementation of two methods: classification and summarization. The classification method is inspired by the training approach of CLIP. It constructs data ${q, a}$ based on the training set $T_{train}$ to train $\mathcal{M}_j$, achieving semantic alignment between questions and answers. The summarization method utilizes the inherent reasoning and summarization capabilities of the large language model itself. By constructing prompts and fine-tuning domain knowledge, $\mathcal{M}_j$ learns to independently weigh the credibility of $\mathcal{M}_i$ and $\mathcal{M}_e$'s answers and respond accordingly.

\subsection{Model-Collaborative Inference}
During the Inference stage, DuetRAG employs a multi-model collaboration approach to obtain answers. For a given question $q \in Q_{test}$, the Internal Knowledge Based Model Mi attempts to provide a direct answer $a_i$. Simultaneously, DuetRAG utilizes a retriever to obtain relevant external knowledge documents $D_q = \{d1, ..., dn\}$, and the External Knowledge Based Model $\mathcal{M}_e$ uses this external knowledge to answer, resulting in answer $a_e$. If the referee model $\mathcal{M}_j$ is implemented using the classification method, $\mathcal{M}_j$ will score the confidence levels of $a_i$ and $a_e$ separately based on question $q$, selecting the answer $a_j$ with the highest score. We note that the External Knowledge Based Model has access to more knowledge, so when there is only a small difference in confidence scores between the two answers, $\mathcal{M}_j$ is designed to lean towards selecting $a_e$. If $\mathcal{M}_j$ is implemented using the summarization method, a prompt constructed based on $q$, $a_i$, and $a_e$ will be provided to $\mathcal{M}_j$, and the answer $a_j$ will be obtained. This prompt contains information such as task instructions and answer preferences.

\section{Evaluation}
We conducted a series of experiments to validate the performance of DuetRAG. Our experiments were based on the LLAMA-7B model \citep{touvron2023llama} and the HotpotQA dataset \citep{yang2018hotpotqa}. For the Ablation Study, we tested the performance of replacing modules in DuetRAG with GPT-3.5 and other methods. In future versions, we plan to evaluate DuetRAG's performance on a wider range of task sets and assess its transferability based on different models.

\subsection{Datasets}
Our experiments were conducted based on the HotpotQA dataset and a collection of external knowledge from Wikipedia. In subsequent versions, we aim to include validation work on additional datasets.

HotpotQA \citep{yang2018hotpotqa} is an open-domain question-answering dataset based on Wikipedia, primarily focused on common knowledge such as movies, sports, history facts, etc. Most of the questions in HotpotQA are in the form of multi-hop questions, requiring multiple steps of reasoning and summarization to arrive at an answer. In our experiments, we used the first 5,000 data samples from the official training set containing 70,000 samples as training data, and the first 1,000 data samples from the official validation set containing 7,000 samples as validation data. The external knowledge documents were sourced from Wikipedia and the fullwiki version provided by the HotpotQA dataset.

\begin{table}
	\caption{Overall Performance of DuetRAG}
	\centering
	\begin{tabular}{llc}
		\toprule
	             & &Accuracy(GPT-4 Eval)     \\
		\midrule
		LLaMA2-7B           & & 12.7   \\
		LLaMA2-7B + RAG     & & 9.2   \\
        DSF(LLaMA2-7B)      & & 20.8         \\
        DSF+RAG(LLaMA2-7B)  & & 25.4 \\
        \midrule
        Internal Model $\mathcal{M}_i$ & & 23.1\\
        External Model $\mathcal{M}_e$ & & 32.7\\
        \textbf{DuetRAG}                & & \textbf{36.3}\\
		\bottomrule
	\end{tabular}
	\label{tab:table}
\end{table}

\subsection{Baselines}
We consider the following baselines for our experiments. All baselines were based on the LLAMA-7B model. In future versions, we will test more base models to validate the transferability of DuetRAG.

\begin{itemize}
    \item \textbf{LLAMA 2-7B model with 0-shot prompting}: this is the commonly used model for QA tasks. We provide clearly written instructions without any reference documents.
	\item \textbf{LLAMA 2-7B model with RAG(Llama2+RAG)}: much similar to the previous setting, expect we include reference documents. This is a popular technique when dealing with domain specific QA tasks.
	\item \textbf{Domain specific fine-tuning with 0-shot prompting(DSF)} : Performing standard supervised fine-tuning without documents in context. Much similar to internal knowledge based model in DuetRAG except DSF is fine-tuned with full trainning data.
	\item \textbf{Domain specific fine-tuning with RAG(DSF+RAG)}: Performing standard supervised fine-tuning without documents in context. Much similar to external knowledge based model in DuetRAG.
\end{itemize}

\subsection{Results}
In Table 1, we validate the performance of DuetRAG based on the aforementioned datasets and baselines. Compared to LLAMA2-7B models based on the 0-shot prompt and RAG, DuetRAG shows nearly a two-fold improvement, demonstrating strong adaptability to domain-specific questions. Additionally, compared to Domain Specific Fine-tuning and RAG Based Domain Specific Fine-tuning, DuetRAG also exhibits significant improvements. Since HotpotQA is a multi-hop question-answering dataset, where retriever results may not always cover all documents containing relevant answer information, the performance of RAG-based models does not significantly surpass 0-shot prompt-based models. Furthermore, compared to Domain Specific Fine-tuned models trained on the entire training set, the performance of our $\mathcal{M}_i$ and $\mathcal{M}_e$ exceeds the former two, indicating potential overfitting of the models on excessive training data. Finally, the performance of DuetRAG surpasses that of $\mathcal{M}_i$ and $\mathcal{M}_e$, demonstrating the effectiveness of the strategy designed for $\mathcal{M}_j$.

\subsection{Ablation Study}

For the Ablation Study, we tested the effectiveness of employing different methods as the referee model $\mathcal{M}_j$, including classification methods based on semantic alignment, text feature-based classification methods, an ensemble method based on 0-shot LLaMA-7B, and an ensemble method utilizing ChatGPT-3.5 (in Table 2).
\begin{table}
	\caption{Performance of $\mathcal{M}_j$ via Multiple Approach}
	\centering
	\begin{tabular}{lcccc}
		\toprule
        \ & \multicolumn{2}{c}{\textbf{Classification}} & \multicolumn{2}{c}{\textbf{Summarization}} \\
        \cmidrule(r){2-3}
        \cmidrule(r){4-5}
	        & Alignment & Text Feature & LLaMA2-7B & ChatGPT-3.5\\
		\midrule
        \textbf{Acc. of $\mathcal{M}_j$} & 36.3 & 34.7 & 33.3 & 39.3 \\
        \textbf{Acc. Compared with $\mathcal{M}_e$} & +3.6 & +2.0 & +0.6 & +6.6 \\
		\bottomrule
	\end{tabular}
	\label{tab:table}
\end{table}
We can see that the classification method based on semantic alignment improved accuracy by 3.6 percent, surpassing the direct text feature-based classification method. Although the Summarization Method using ChatGPT-3.5 demonstrated significant performance enhancements, these improvements are primarily attributed to the powerful summarization capabilities of ChatGPT due to its large-scale training. However, similar summarization methods applied to the 0-shot LLaMA2-7B model did not yield particularly favorable results, likely due to the limited summarization capabilities of LLaMA2-7B itself.


\section{Conclusion}
DUET is a novel multi-model collaboration framework designed to enhance the performance of question-answering models in specific domains with few training samples and complex environments. We have identified several key design decisions, such as simultaneously using an Internal Knowledge-Based Model and an External Knowledge-Based Model to generate answers, and using a third model to generate the answer with the highest confidence level. Our evaluation on HotpotQA highlights the significant potential of DUET. Looking ahead, we anticipate that Retrieval-Augmented Generation (RAG) within the domain will continue to attract attention in both industry and academia. Unlike general RAG, our work addresses the practical scenario where LLMs are tasked with answering questions using domain-specific knowledge. Meanwhile, we will continue to update this work and validate the reliability and transferability of DUET on more application scenarios and base models.

\bibliographystyle{unsrtnat}
\bibliography{references}  

\begin{thebibliography}{20}
\providecommand{\natexlab}[1]{#1}
\providecommand{\url}[1]{\texttt{#1}}
\expandafter\ifx\csname urlstyle\endcsname\relax
  \providecommand{\doi}[1]{doi: #1}\else
  \providecommand{\doi}{doi: \begingroup \urlstyle{rm}\Url}\fi

\bibitem[Achiam et~al.(2023)Achiam, Adler, Agarwal, Ahmad, Akkaya, Aleman, Almeida, Altenschmidt, Altman, Anadkat, et~al.]{achiam2023gpt}
Josh Achiam, Steven Adler, Sandhini Agarwal, Lama Ahmad, Ilge Akkaya, Florencia~Leoni Aleman, Diogo Almeida, Janko Altenschmidt, Sam Altman, Shyamal Anadkat, et~al.
\newblock Gpt-4 technical report.
\newblock \emph{arXiv preprint arXiv:2303.08774}, 2023.

\bibitem[Wei et~al.(2022)Wei, Wang, Schuurmans, Bosma, Xia, Chi, Le, Zhou, et~al.]{wei2022chain}
Jason Wei, Xuezhi Wang, Dale Schuurmans, Maarten Bosma, Fei Xia, Ed~Chi, Quoc~V Le, Denny Zhou, et~al.
\newblock Chain-of-thought prompting elicits reasoning in large language models.
\newblock \emph{Advances in neural information processing systems}, 35:\penalty0 24824--24837, 2022.

\bibitem[Kirkpatrick et~al.(2017)Kirkpatrick, Pascanu, Rabinowitz, Veness, Desjardins, Rusu, Milan, Quan, Ramalho, Grabska-Barwinska, et~al.]{kirkpatrick2017overcoming}
James Kirkpatrick, Razvan Pascanu, Neil Rabinowitz, Joel Veness, Guillaume Desjardins, Andrei~A Rusu, Kieran Milan, John Quan, Tiago Ramalho, Agnieszka Grabska-Barwinska, et~al.
\newblock Overcoming catastrophic forgetting in neural networks.
\newblock \emph{Proceedings of the national academy of sciences}, 114\penalty0 (13):\penalty0 3521--3526, 2017.

\bibitem[Lewis et~al.(2020)Lewis, Perez, Piktus, Petroni, Karpukhin, Goyal, K{\"u}ttler, Lewis, Yih, Rockt{\"a}schel, et~al.]{lewis2020retrieval}
Patrick Lewis, Ethan Perez, Aleksandra Piktus, Fabio Petroni, Vladimir Karpukhin, Naman Goyal, Heinrich K{\"u}ttler, Mike Lewis, Wen-tau Yih, Tim Rockt{\"a}schel, et~al.
\newblock Retrieval-augmented generation for knowledge-intensive nlp tasks.
\newblock \emph{Advances in Neural Information Processing Systems}, 33:\penalty0 9459--9474, 2020.

\bibitem[Zhang et~al.(2023)Zhang, Li, Cui, Cai, Liu, Fu, Huang, Zhao, Zhang, Chen, et~al.]{zhang2023siren}
Yue Zhang, Yafu Li, Leyang Cui, Deng Cai, Lemao Liu, Tingchen Fu, Xinting Huang, Enbo Zhao, Yu~Zhang, Yulong Chen, et~al.
\newblock Siren's song in the ai ocean: a survey on hallucination in large language models.
\newblock \emph{arXiv preprint arXiv:2309.01219}, 2023.

\bibitem[Yang et~al.(2018)Yang, Qi, Zhang, Bengio, Cohen, Salakhutdinov, and Manning]{yang2018hotpotqa}
Zhilin Yang, Peng Qi, Saizheng Zhang, Yoshua Bengio, William~W Cohen, Ruslan Salakhutdinov, and Christopher~D Manning.
\newblock Hotpotqa: A dataset for diverse, explainable multi-hop question answering.
\newblock \emph{arXiv preprint arXiv:1809.09600}, 2018.

\bibitem[Guu et~al.(2020)Guu, Lee, Tung, Pasupat, and Chang]{guu2020retrieval}
Kelvin Guu, Kenton Lee, Zora Tung, Panupong Pasupat, and Mingwei Chang.
\newblock Retrieval augmented language model pre-training.
\newblock In \emph{International conference on machine learning}, pages 3929--3938. PMLR, 2020.

\bibitem[Borgeaud et~al.(2022)Borgeaud, Mensch, Hoffmann, Cai, Rutherford, Millican, Van Den~Driessche, Lespiau, Damoc, Clark, et~al.]{borgeaud2022improving}
Sebastian Borgeaud, Arthur Mensch, Jordan Hoffmann, Trevor Cai, Eliza Rutherford, Katie Millican, George~Bm Van Den~Driessche, Jean-Baptiste Lespiau, Bogdan Damoc, Aidan Clark, et~al.
\newblock Improving language models by retrieving from trillions of tokens.
\newblock In \emph{International conference on machine learning}, pages 2206--2240. PMLR, 2022.

\bibitem[Shi et~al.(2023)Shi, Min, Yasunaga, Seo, James, Lewis, Zettlemoyer, and Yih]{shi2023replug}
Weijia Shi, Sewon Min, Michihiro Yasunaga, Minjoon Seo, Rich James, Mike Lewis, Luke Zettlemoyer, and Wen-tau Yih.
\newblock Replug: Retrieval-augmented black-box language models.
\newblock \emph{arXiv preprint arXiv:2301.12652}, 2023.

\bibitem[Lin et~al.(2023)Lin, Chen, Chen, Shi, Lomeli, James, Rodriguez, Kahn, Szilvasy, Lewis, et~al.]{lin2023ra}
Xi~Victoria Lin, Xilun Chen, Mingda Chen, Weijia Shi, Maria Lomeli, Rich James, Pedro Rodriguez, Jacob Kahn, Gergely Szilvasy, Mike Lewis, et~al.
\newblock Ra-dit: Retrieval-augmented dual instruction tuning.
\newblock \emph{arXiv preprint arXiv:2310.01352}, 2023.

\bibitem[Izacard et~al.(2022)Izacard, Lewis, Lomeli, Hosseini, Petroni, Schick, Dwivedi-Yu, Joulin, Riedel, and Grave]{izacard2022atlas}
Gautier Izacard, Patrick Lewis, Maria Lomeli, Lucas Hosseini, Fabio Petroni, Timo Schick, Jane Dwivedi-Yu, Armand Joulin, Sebastian Riedel, and Edouard Grave.
\newblock Atlas: Few-shot learning with retrieval augmented language models.
\newblock \emph{arXiv preprint arXiv:2208.03299}, 2022.

\bibitem[Zhang et~al.(2024)Zhang, Patil, Jain, Shen, Zaharia, Stoica, and Gonzalez]{zhang2024raft}
Tianjun Zhang, Shishir~G Patil, Naman Jain, Sheng Shen, Matei Zaharia, Ion Stoica, and Joseph~E Gonzalez.
\newblock Raft: Adapting language model to domain specific rag.
\newblock \emph{arXiv preprint arXiv:2403.10131}, 2024.

\bibitem[Wang et~al.(2023)Wang, Ping, McAfee, Xu, Li, Shoeybi, and Catanzaro]{wang2023instructretro}
Boxin Wang, Wei Ping, Lawrence McAfee, Peng Xu, Bo~Li, Mohammad Shoeybi, and Bryan Catanzaro.
\newblock Instructretro: Instruction tuning post retrieval-augmented pretraining.
\newblock \emph{arXiv preprint arXiv:2310.07713}, 2023.

\bibitem[Liu et~al.(2024)Liu, Ping, Roy, Xu, Shoeybi, and Catanzaro]{liu2024chatqa}
Zihan Liu, Wei Ping, Rajarshi Roy, Peng Xu, Mohammad Shoeybi, and Bryan Catanzaro.
\newblock Chatqa: Building gpt-4 level conversational qa models.
\newblock \emph{arXiv preprint arXiv:2401.10225}, 2024.

\bibitem[Li et~al.(2024{\natexlab{a}})Li, Hammoud, Itani, Khizbullin, and Ghanem]{li2024camel}
Guohao Li, Hasan Hammoud, Hani Itani, Dmitrii Khizbullin, and Bernard Ghanem.
\newblock Camel: Communicative agents for" mind" exploration of large language model society.
\newblock \emph{Advances in Neural Information Processing Systems}, 36, 2024{\natexlab{a}}.

\bibitem[Du et~al.(2023)Du, Li, Torralba, Tenenbaum, and Mordatch]{du2023improving}
Yilun Du, Shuang Li, Antonio Torralba, Joshua~B Tenenbaum, and Igor Mordatch.
\newblock Improving factuality and reasoning in language models through multiagent debate.
\newblock \emph{arXiv preprint arXiv:2305.14325}, 2023.

\bibitem[Liang et~al.(2023)Liang, He, Jiao, Wang, Wang, Wang, Yang, Tu, and Shi]{liang2023encouraging}
Tian Liang, Zhiwei He, Wenxiang Jiao, Xing Wang, Yan Wang, Rui Wang, Yujiu Yang, Zhaopeng Tu, and Shuming Shi.
\newblock Encouraging divergent thinking in large language models through multi-agent debate.
\newblock \emph{arXiv preprint arXiv:2305.19118}, 2023.

\bibitem[Sun et~al.(2023)Sun, Yin, Li, Wu, Qiu, and Kong]{sun2023corex}
Qiushi Sun, Zhangyue Yin, Xiang Li, Zhiyong Wu, Xipeng Qiu, and Lingpeng Kong.
\newblock Corex: Pushing the boundaries of complex reasoning through multi-model collaboration.
\newblock \emph{arXiv preprint arXiv:2310.00280}, 2023.

\bibitem[Li et~al.(2024{\natexlab{b}})Li, Peng, He, Galley, Gao, and Yan]{li2024guiding}
Zekun Li, Baolin Peng, Pengcheng He, Michel Galley, Jianfeng Gao, and Xifeng Yan.
\newblock Guiding large language models via directional stimulus prompting.
\newblock \emph{Advances in Neural Information Processing Systems}, 36, 2024{\natexlab{b}}.

\bibitem[Touvron et~al.(2023)Touvron, Martin, Stone, Albert, Almahairi, Babaei, Bashlykov, Batra, Bhargava, Bhosale, et~al.]{touvron2023llama}
Hugo Touvron, Louis Martin, Kevin Stone, Peter Albert, Amjad Almahairi, Yasmine Babaei, Nikolay Bashlykov, Soumya Batra, Prajjwal Bhargava, Shruti Bhosale, et~al.
\newblock Llama 2: Open foundation and fine-tuned chat models.
\newblock \emph{arXiv preprint arXiv:2307.09288}, 2023.

\end{thebibliography}






\end{document}